\documentclass[11pt,a4paper]{article}
\usepackage[hyperref]{eacl2021}
\usepackage{times}
\usepackage{latexsym}

\usepackage{microtype}

\aclfinalcopy 

\usepackage{url}
\usepackage{amsmath,amssymb}
\DeclareMathOperator{\E}{\mathbb{E}}
\usepackage{amsfonts}
\usepackage{helvet}
\usepackage{courier}
\usepackage{url}
\usepackage{float,color}
\usepackage{times}
\usepackage{algorithm}
\usepackage{algorithmic}
\usepackage{epsfig}
\usepackage{stfloats}
\usepackage{url}
\usepackage{diagbox}
\usepackage{subfigure}
\usepackage{epstopdf}
\usepackage{multicol}
\usepackage{dsfont,amsfonts,amssymb,amsmath,color}
\usepackage{multirow}
\usepackage{booktabs}
\def\0{{\bf 0}}
\def\1{{\bf 1}}

\def\PM{{\mathcal P}}
\def\JM{{\mathcal J}}
\def\LM{{\mathcal L}}
\def\RM{{\mathcal R}}

\def\NM{{\mathcal N}}
\def\OM{{\mathcal O}}
\def\SM{{\mathcal S}}

\title{Variational Weakly Supervised Sentiment Analysis with \\
Posterior Regularization}

\author{Ziqian Zeng, Yangqiu Song \\
  Department of CSE, HKUST, Hong Kong, China\\
   {\tt \{zzengae, yqsong\}@cse.ust.hk } \\
}

\date{}

\begin{document}
\maketitle

\begin{abstract}
Sentiment analysis is an important task in natural language processing (NLP). 
Most of existing state-of-the-art methods are under the supervised learning paradigm.
However, human annotations can be scarce. 
Thus, we should leverage more weak supervision for sentiment analysis.
In this paper, we propose a posterior regularization framework for the variational approach to the weakly supervised sentiment analysis to better control the posterior distribution of the label assignment. 
The intuition behind the posterior regularization is that if extracted opinion words from two documents are semantically similar, the posterior distributions of two documents should be similar. 
Our experimental results show that the posterior regularization can improve the original variational approach to the weakly supervised sentiment analysis and the performance is more stable with smaller prediction variance.
\end{abstract}
\section{Introduction}
Sentiment analysis is a task of identifying the sentiment polarity expressed in textual data \cite{liu2012sentiment}. 
Most state-of-the-art sentiment analysis methods in the literature are supervised methods which require many labeled training data. 
However, human annotations in the real world are scarce. 
While we assume there is abundant annotated data to train more and more complex models, 
there is still a need to consider weakly supervised methods that require less human annotation. 

One way to perform weakly supervised sentiment analysis is using a predefinited lexicon \cite{turney2002thumbs,maite2011lexicon}. 
A lexicon consists of many opinion words. 
For each opinion word, its polarity (positive or negative) and strength (the degree to which the opinion word is positive or negative) are annotated by domain experts. 
lexicon-based weakly supervised methods perform a dictionary lookup and assign a polarity according to all opinion words extracted from a document. 
A good lexicon requires high precision and high coverage, which needs a lot of human effort.

Another way to do weakly supervised sentiment analysis is using limited keywords \cite{meng2018weakly,zeng2019variational}. 
Compared with lexicon-based methods, user-provided keywords require less human effort. 
Among keyword-based methods, there are two directions. 
First, \cite{meng2018weakly} leveraged limited keywords to expand more keywords and generate pseudo-labeled data, and then performed self-training on real unlabeled data for model refinement. 
Possible improvements of this direction include investigating more advanced keywords expansion techniques to generate better pseudo-labeled samples~\cite{miller2012using} and developing more advanced self-training algorithms\cite{coden2014semantic}. 

Second, the Variational Weakly Supervised (VWS) sentiment analysis \cite{zeng2019variational} used target-opinion word pairs as supervision signal. 
Its objective function is to predict an opinion word given a target word. 
For example, in a sentence ``the room is big,'' ``room'' is a target word and ``big'' is an opinion word. 
By introducing a latent variable (the sentiment polarity), they can learn a well-approximated posterior distribution via optimizing the evidence lower bound. 
The posterior probability here is the probability of a possible polarity (e.g., positive or negative) given a document, which is a typical sentiment classifier.

A potential issue with VWS is that optimizing the objective function may not guide the role of the latent variable to be sentiment polarity. 
For example, when half of reviews mention ``big room'' and half of reviews mention ``small room,'' the latent variable is possibly related to the size of rooms, but the expected role of the latent variable is the sentiment polarity of rooms. 
Hence how to control and regularize the posterior distribution is very important. 
One indirect way to control the posterior distribution is clever initialization \cite{ganchev2010posterior}. 
Originally, VWS aims to predict the sentiment polarity of each aspect for the multi-aspect sentiment analysis. 
So it uses the overall sentiment polarity to pretrain the model so that the posterior distribution has a good initialization. 
The overall polarity and polarity of each aspect are highly correlated. 
Thus, the initialization is highly likely to be similar to the true posterior distribution.

In this paper, we propose to use posterior regularization to regularize the VWS approach for sentiment analysis. 
There are two types of side information we can leverage to regularize the latent variable. 
First, calculating the similarity between keywords and extracted opinion words from a document can guide the model to decide which polarity that the document belongs to. 
Second, calculating the similarity between extracted opinion words from two documents can guide the model to decide whether two documents belong to the same polarity. 
The first type of side information is very easy to leverage, and it is reflected by our pretraining process. 
When a document is similar to keywords associated with a specific polarity, then we enforce that the posterior probability of a specific polarity should be large. 
In this case, in the pretraining process, we assign a pseudo label to the document. 
The second type of side information does not directly suggest which sentiment polarity that a document should be assigned to.  
It enforces pairwise constraints to the model. 
Our proposed posterior regularization leverage the second side information to ensure that when two documents are similar (dissimilar), the regularization enforces the posterior distribution of two documents to be similar (dissimilar). 

Our contributions are summarized as follows,

$\bullet$ We develop a posterior regularization framework for the variational weakly supervised sentiment analysis. 

$\bullet$ The experimental results show that the proposed regularization can improve the VWS model, make the results more stable, and outperform other weakly supervised baselines.

Our code is available at \url{https://github.com/HKUST-KnowComp/VWS-PR}.

\section{Methodology}
In this section, we first review the variational weakly supervised (VWS) sentiment analysis method in Section \ref{sec:vws}. Then we introduce our posterior regularization in Section \ref{sec:pr}.

\subsection{VWS Sentiment Analysis}\label{sec:vws}
Before formally introducing the VWS framwork, we give a concrete example to illustrate how VWS works.
Let $\mathbf{x}$ be the representation of a document $x$, e.g., bag of words or feature outputs of neural networks. 
Let $C$ be a random variable, indicating the sentiment polarity of a document. 
The possible value assignment of $C$ can be \textit{positive} or \textit{negative}, or rating from $1$ to $10$. 
Suppose there is a document $x$ where we extract an opinion word ``terrific.'' 
The objective function is to maximize the probability of opinion word ``terrific.'' 
By introducing a latent variable $C$, the objective function is split into two probabilities, corresponding to two classifiers, namely, sentiment polarity classifier and opinion word classifier. 
The input of sentiment classifier is the document representation $\mathbf{x}$, and it produces a probability distribution of sentiment polarity, i.e., $p(C=positive|\mathbf{x})$ and $p(C=negative|\mathbf{x})$.
The input of opinion word classifier is extracted opinion words and the estimated sentiment polarity distribution, 
and it produces a probability distribution of opinion word given estimated sentiment polarity distribution, i.e., $p(\text{``terrific''}|C=positive)$ and $p(\text{``terrific''}|C=negative)$. 
%
%
%
%
%


\subsubsection{Sentiment Polarity Classifier}
The sentiment polarity classifier aims to estimate a distribution $q(C|\mathbf{x})$, where $C$ is a discrete random variable representing the sentiment polarity of a document. 
Let $c$ denote a possible value of the random variable $C$, representing a possible value of sentiment polarity, e.g., \textit{positive} or \textit{negative}. 
The sentiment classifier estimates the probability as
\begin{equation}\label{eq:encoder-softmax}
q(C = c|\mathbf{x}) = \frac{\exp \big( \mathbf{w}_{c}^{T} \mathbf{x} \big)}{\sum_{c'}{\exp \big( \mathbf{w}_{c'}^{T} \mathbf{x} \big)}} \; ,
\end{equation}
\begin{equation}
\mathbf{x} = CNN(x) \;,
\end{equation}
where $\mathbf{w}_{c}$ is a trainable vector associated with a sentiment polarity $c$, $x$ is a document, and $\mathbf{x}$ is the document representation.  
The representation of a document $\mathbf{x}$ can be various. We use Convolutional Neural Network (CNN) in the experiment. 

\subsubsection{Opinion Word Classifier}
The opinion word classifier aims to estimate the probability of an opinion word $w_o$ given a possible value of sentiment polarity $c$:
\begin{equation}
p(w_o |c) = \frac{\exp \big( \varphi ({w_o}, c) \big)}{ \sum_{{w_o'}}{ \exp \big( \varphi ({w_o'}, c) \big) } } \;,
\label{oc}
\end{equation}
where $\varphi(\cdot)$ is a scoring function taking opinion word $w_o$ and a possible value of sentiment polarity $c$ as inputs. 
The nature of the scoring function is about the frequency of occurrence.  
If an opinion word and a possible value of sentiment polarity co-occur frequently, the score will be high, otherwise, it will be low. Specifically, we define:
\begin{equation}
\varphi ({w_o}, c)=  \mathbf{a}_{c}^T \mathbf{w}_o \;, \label{score}
\end{equation}
where $\mathbf{w}_o$ is the trainable word embedding of opinion word $w_o$, $\mathbf{a}_{c}$ is a trainable vector associated with $c$. 
The scoring function can be various, e.g., multilayer perceptron (MLP).
Here we only introduce the simplest case. 

Given a possible value of sentiment polarity $c$, VWS aims to maximize the probability of opinion words that frequently occurred with $c$. 
For example, the opinion word ``good'' is usually occurred with sentiment polarity \textit{positive}, and the opinion word ``terrible'' is usually occurred with sentiment polarity \textit{negative}. 

\subsubsection{Training Objective}
The objective function of VWS is to maximize the log-likelihood of an opinion word $w_o$. 
After introducing a latent variable (i.e., the sentiment polarity of a document) to the objective function, we can derive a variational lower bound of the log-likelihood which can incorporate two classifiers. 
The first one corresponds to the sentiment classifier.
The second one corresponds to the opinion word classifier. 
The variational lower bound of log-likelihood is shown as follows:
\begin{align}
\LM_{1} & = \sum_{x \in X} \sum_{w_o \in \PM_{x}} \log p(w_o) \nonumber \\
& = \sum_{x \in X} \sum_{w_o \in \PM_{x}} \log \sum_{c}p(w_o , c) \nonumber\\
& = \sum_{x \in X} \sum_{w_o \in \PM_{x}} \log \sum_{c} q(c|\mathbf{x}) \Big[ \frac{p(w_o , c )}{q(c|\mathbf{x})} \Big] \nonumber \\
& \geq \sum_{x \in X} \sum_{w_o \in \PM_{x}} \sum_{c} q(c|\mathbf{x}) \Big[ \log \frac{p(w_o , c)}{q(c|\mathbf{x})} \Big] \nonumber \\
& = \sum_{x \in X} \sum_{w_o \in \PM_{x}} \E_{q(C|\mathbf{x})} \big[\log p(w_o|c)p(c) \big]  \nonumber \\ 
& + \sum_{x \in X} \sum_{w_o \in \PM_{x}} H(q(C|\mathbf{x})), \label{vari}
\end{align}
where $X$ is the training set containing all documents, and $\PM_{x}$ is the set of all opinion words extracted from a document $x$, $H(\cdot)$ refers to the Shannon entropy, and $q(c|\mathbf{x})$ is short for $q(C=c|\mathbf{x})$. 
By applying Jensen's inequality, the log-likelihood is lower-bounded by Eq. (\ref{vari}). 
The equality holds if and only if the KL-divergence of two distributions, $q(C|\mathbf{x})$ and $p(C|w_o)$, equals to zero. 
Maximizing the evidence lower bound is equivalent to minimizing the KL-divergence. 
Hence, VWS can learn a sentiment classifier that can produce a similar distribution to the true posterior $p(C|w_o)$. 
We assume that the training set is perfectly balanced, which means the prior distribution of sentiment polarity, i.e., $p(C)$, is a uniform distribution. 
Hence, $p(c)$ is a constant, which can be ignored. 

\subsubsection{Approximation}
The partition function in Eq. (\ref{oc}) requires the summation over all opinion words in the vocabulary. 
Since the size of the opinion word vocabulary is large, VWS uses the negative sampling technique \cite{mikolov2013distributed} to approximate Eq. (\ref{oc}).
Specifically, VWS approximates $p(w_o|c)$ in the objective (\ref{oc}) with the following objective function:
\begin{equation}
\log \sigma \big( \varphi ({w_o}, c) \big) + \sum_{ w_o' \in \NM } \log\big( 1 - \sigma \big(\varphi ({w_o'}, c) \big) \big) \; \label{app1},
\end{equation}
where $w_o'$ is a negative sample in opinion words vocabulary, $\NM$ is the set of negative samples and $\sigma(\cdot)$ is the sigmoid function. 
In order to ensure that the approximation part and the entropy term are on the same scale~\cite{marcheggiani2016discrete}, a hyper-parameter $\alpha$ is added to the entropy term. 
The objective function becomes:
\begin{align} 
\LM_2 & = \sum_{x \in X} \sum_{w_o \in \PM_{x}}  \E_{q(C|\mathbf{x})} \big[ \log \sigma \big( \varphi ({w_o}, c) \big) \nonumber \\
\quad \quad & + \sum_{ w_o' \in \NM } \log \big( 1 - \sigma \big( \varphi ({w_o'}, c) \big) \big) + \log p(c) \big] \nonumber \\
\quad \quad &+\sum_{x \in X} \sum_{w_o \in \PM_{x}} \alpha H(q(C|\mathbf{x}))  \label{obj_w_app1}.
\end{align}

\subsection{Posterior Regularization} \label{sec:pr}
As pointed out by \cite{ganchev2010posterior}, controlling the posterior distribution is crucial for models that estimate posterior distribution by maximizing the likelihood of the observed data via marginalizing over the latent variables. 
We need side information to regularize the posterior distribution. 
The side information we leveraged is that if the opinion words extracted from two documents are similar semantically, then these two documents probably are in the same class, and if the opinion words are opposite semantically, then these two documents are probably not in the same class. 
For example, if one document $x^i$ contains opinion words ``great'' and ``awesome,'' another document $x^j$ contains opinion words ``great'' and ``excellent,'' and another document $x^k$ contains opinion words ``awful'' and ``terrible,'' it is highly possible that $x^i$ and $x^j$ belong to the same class because their extracted opinion words are similar semantically, and $x^i$ and $x^k$ do not belong to the same class because their extracted opinion words are opposite semantically. 

We formulate our posterior regularization as:
\begin{align}\label{pr}
  \RM(x^i,x^j) &= -d(x^i,x^j) \cdot \SM(\OM(x^i),\OM(x^j)),
\end{align}
where $d(x^i,x^j)$ is short for $d(q(C|\mathbf{x}^i),q(C|\mathbf{x}^j))$, meaning the distance of two posterior distributions, and we use Euclidean distance metric; $\SM(\cdot,\cdot)$ is a score function which measures the similarity or dissimilarity between two sets of opinion words; $\OM(x^i)$ represents all opinion words extracted from a document $x^i$. 
We finally maximize Eq. (\ref{pr}) in the objective function. 
When $\SM(\OM(x^i),\OM(x^j))$ is positive (suggesting similar), this regularization enforces the distance to be small, and when $\SM(\OM(x^i),\OM(x^j))$ is negative (suggesting dissimilar), this regularization enforces the distance to be large. 
When $\SM(\OM(x^i),\OM(x^j))$ is zero, it suggests comparison between opinion words cannot decide whether two documents are similar or not.

Next, we will introduce the scoring function $\SM(\cdot,\cdot)$. Suppose a document $x^i$ contains a set of opinion words $\OM(x^i) = \{w^i_{o_1},w^i_{o_2},\cdots,w^i_{o_k}\}$ and a document $x^j$ contains a set of opinion words $\OM(x^j) = \{w^j_{o_1},w^j_{o_2},\cdots,w^j_{o_k}\}$. 
We define an operation $cos(\OM(x^i),\OM(x^j))$ over two sets of opinion words. 
It will return all cosine similarity values of all valid opinion word pairs where one word must come from $\OM(x^i)$ and the other must come from $\OM(x^j)$. 
We represent opinion words using embeddings, i.e., the embeddings in the opinion word classifier in Eq. (\ref{score}). 
If there are $k$ opinion words in each set,  $cos(\OM(x^i),\OM(x^j))$ will return $\frac{k*(k-1)}{2}$ cosine similarity values. 
When we want to know whether two documents are similar in opinion words, we pay attention to the maximum value, i.e., $max\_cos = max\big(cos(\OM(x^i),\OM(x^j))\big)$. 
When we want to know whether two documents are dissimilar in opinion words, we pay attention to the minimum value, i.e., $min\_cos = min\big(cos(\OM(x^i),\OM(x^j))\big)$. 
So we define:
\begin{equation}
\SM(\cdot,\cdot) =
\begin{cases}

\multirow{2}{*}{${max\_cos}$,} & max\_cos > \gamma_1 \quad \multirow{2}{*}{$\&$}\\
                               & min\_cos \geq \gamma_2,\vspace{0.1in} \\
                             
\multirow{2}{*}{${min\_cos}$,} & max\_cos \leq \gamma_1 \quad  \multirow{2}{*}{$\&$} \\
                            & min\_cos < \gamma_2, \vspace{0.1in}\\

\multirow{2}{*}{$\delta$,} & max\_cos > \gamma_1 \quad \multirow{2}{*}{$\&$} \\
                            & min\_cos < \gamma_2, \vspace{0.1in}\\
0, & otherwise,
\end{cases}
\end{equation}
where $\SM(\cdot,\cdot)$ is short for $\SM((\OM(x^i),\OM(x^j))$ due to space limit. 
The first condition means two documents have some semantically similar opinion words ($max\_cos > \gamma_1$) and have no semantically dissimilar opinion words ($min\_cos \geq \gamma_2$). The value returned by the function score is $max\_cos$. It should be a positive value. 
The second condition means two documents have no semantically similar opinion words ($max\_cos \leq \gamma_1$) and have some semantically dissimilar opinion words ($min\_cos < \gamma_2$). The value returned by function score is $min\_cos$. It should be a negative number. 
The third condition means two documents have some semantically similar opinion words ($max\_cos > \gamma_1$) and also have some semantically dissimilar opinion words ($min\_cos < \gamma_2$). 
This condition corresponds to a real-world situation that when some customers want to express some negative sentiment, they usually point out some positive aspects first, and then start with a ``but'', and emphasize some negative aspects. 
The opinion words sets extracted from these type of documents have both negative and positive opinion words. 
When we compare two of them, they will have some similar opinion words and dissimilar opinion words. 
In this case, we tend to assume they are in the same class. If the third condition is satisfied, it will return an non-negative value $\delta \in [0,1]$; 
The final condition means two documents have no semantically similar opinion words ($max\_cos \leq \gamma_1$) and have no semantically dissimilar opinion words ($min\_cos \geq \gamma_2$). It will return $0$.

The mechanism of the regularization is that if the posterior distributions $q(C|\mathbf{x}^i)$ and $q(C|\mathbf{x}^j)$ are different from each other, i.e., $d\big( q(C|\mathbf{x}^i), q(C|\mathbf{x}^j) \big)$ is large, but opinion words suggest that these two documents should be in the same cluster i.e., $s(w^{i}_{d}, w^{j}_{d})$ is large, then $d\big( q(C|\mathbf{x}^i), q(C|\mathbf{x}^j) \big)$ will be encouraged to be small by applying the regularization. 
Oppositely, if the posterior distributions $q(C|\mathbf{x}^i)$ and $q(C|\mathbf{x}^j)$ are similar, i.e., $d\big( q(C|\mathbf{x}^i), q(C|\mathbf{x}^j) \big)$ is small, but opinion words suggest that these two documents should be in the different cluster i.e., $s(w^{i}_{d}, w^{j}_{d})$ is small, then $d\big( q(C|\mathbf{x}^i), q(C|\mathbf{x}^j) \big)$ will be encouraged to be large by applying the regularization.

The final objective function with posterior regularization is as follows,
\begin{align} \label{final-ob}
\JM & = \LM_{2} + \beta \sum_{x^i \in X} \sum_{x^j \in X} R(x^i,x^j)\;,
\end{align}
where $x^i$ and $x^j$ are documents in the training set $X$.
The constraints are defined in a $|X| \times |X|$ space. 
In practice, we train our model batch by batch. 
So we only apply the constraints within a mini-batch. 
There are at most $|X_b| \times |X_b|$ constraints in a mini-batch, where $|X_b|$ is the number of samples in a mini-batch. 

\section{Experiments}
In this section, we evaluate the empirical performance of our method on binary sentiment classification tasks. 
\subsection{Datasets}
\begin{table*}[t!]
  \centering
  
  \begin{tabular}{l|l|l|l}
  \toprule
  Datasets&
     Yelp & IMDB & Amazon \\
\midrule[1pt]
  Size of training set & 38,000 & 20,000 & 20,000 \\
  Size of development set & 3,800 & 2,000 & 2,000 \\
  Average doc length & 155.12 & 252.90 & 88.28  \\
  Opinion vocabulary size & 1,097 & 688 & 394\\
  \bottomrule
  \end{tabular}
  \caption{Statistics of Yelp, IMDB, and Amazon dataset.}
  \label{tab:stat}
\end{table*}
We use three corpora to evaluate the performance of our proposed method. 
All corpora have two classes and perfectly balanced. 
For all methods, we use a development set for hyper-parameter tuning. 
For all methods, we use the training set as the test set since all methods do not use the ground truth in the training set.

(1) \textbf{Yelp} Review: We use the Yelp reviews polarity dataset from \cite{zhang2015character} and take its test set containing 38,000 documents as the corpus for evaluation. 
For hyper-parameter tuning, we also extract 3,800 documents from the original training set of \cite{zhang2015character} to serve as a development set. 

(2) \textbf{IMDB} Review: We use the IMDB reviews polarity dataset from \cite{maas2011learning} and randomly extract $20,000$ reviews from its original test set as the corpus for evaluation. 
For hyper-parameter tuning, we also extract $2,000$ documents from the original training set of \cite{maas2011learning} to serve as a development set. 

(3) \textbf{Amazon} Review: We use the Amazon reviews polarity dataset from \cite{zhang2015character} and randomly extracted $20,000$ reviews from its original test set as the corpus for evaluation. 
For hyper-parameter tuning, we also extract $2,000$ documents from the original training set of \cite{zhang2015character} to serve as a development set.

Table \ref{tab:stat} provides the details of these datasets.

\subsection{Compared Methods}
\noindent\textbf{Lexicon} uses an opinion lexicon to assign sentiment polarity to a document \cite{read2009weakly,pablos2015v3}.
We combine two popular opinion lexicons used by \cite{hu2004mining} and \cite{wilson2005recognizing} to get a larger lexicon. 
If an extracted opinion is in the positive (negative) lexicon, it votes for positive (negative).
When the opinion word is with a negation word such as ``no'' and ``not'', its polarity will be the opposite.
Then, the polarity of a document is determined by using majority voting among all extracted opinion words.
When the number of positive and negative words is equal, the document will be randomly assigned a polarity.

\noindent\textbf{WeSTClass} \cite{meng2018weakly} first generates pseudo labels for documents which contain user-provided keywords. Keywords are expanded to generate more pseudo samples. It pretrains a CNN/LSTM model using pseudo samples as the training set and then performs a self-training process. Here, we use CNN because it empirically outperforms LSTM. The CNN architecture we used here is the same as the one described in \cite{meng2018weakly}.

\noindent\textbf{Keyword Pretrain} generates pseudo labels for documents in training set which contain user-provided keywords. We pretrain a CNN model using pseudo samples as the training set. In the CNN model, four different filter sizes $\{2, 3, 4, 5\}$ are applied, and a max-pooling layer is applied to each convolutional layer, and each convolutional layer has $100$ filters.

\noindent\textbf{VWS} \cite{zeng2019variational} uses target opinion word pairs as supervision signal. It trains a sentiment polarity classifier and opinion word classifier simultaneously via optimizing the variational lower bound. 
We use a CNN model as the sentiment polarity classifier. 
And we pretrain it by generating pseudo labels for documents which contain user-provided keywords in the training set. 
The CNN architecture is the same as the one in \textbf{Keyword Pretrain}. 

\noindent\textbf{VWS-PR} is VWS method with proposed posterior regularization. 

\subsection{Keywords and Opinion Word Extraction}
\begin{table*}[t!]
  \centering
  
  \begin{tabular}{l|lll|lll}
  \toprule
  Datasets&
     \multicolumn{3}{c}{Positive} & 
     \multicolumn{3}{|c}{Negative} \\
\midrule
  Yelp &  terrific & amazing & awesome & 
horrible & worst & bad  \\
  IMDB & great & fantastic & awesome & awful & worst & bad
  \\
  Amazon & great & fantastic & awesome & poor & worst & bad \\
  \bottomrule
  \end{tabular}
  \caption{Keywords of Yelp, IMDB, and Amazon datasets.}
  \label{tab:keyword}
\end{table*}
We manually select three keywords for each class. 
The details of keywords of three datasets are shown in table \ref{tab:keyword}. 

For opinion word extraction, we adopt four rules proposed by VWS \cite{zeng2019variational} in the implementation. 
All rules rely on dependency parser \cite{chen2014fast}. 
When a target word and an opinion word satisfy a dependency relation, we will extract the opinion word. 
The details of dependency relation and examples are provided in Table \ref{tab:extraction}. 
When a pair of words satisfy one rule, there are still some restrictions on head and tails words to be satisfied. 
There is no restriction for Rule $1$. 
For Rule $2$, the head word should be an adjective and the tail word should a noun. For Rule $3$, the head word should be one of the following four words: ``like,'' ``dislike,'' ``love,'' and ``hate.'' 
For Rule $4$, the head word should be one of the following word: ``seem,'' ``look,'' ``feel,'' ``smell,'' and ``taste.''

\begin{table*}[t!]
  \centering
  \begin{tabular}{l|c|c|c}
  \toprule
  Rule & Dependency Relation & Example & Extracted Word\\
  \midrule
  1 & adjectival modifier & they have delicious food & delicious\\
  2 & nominal subject & the room is big & big \\
  3 & direct object & i like it & like\\
  4 & open clausal complement & i feel comfortable & comfortable\\
  \bottomrule
  \end{tabular}
  \caption{Opinion words extraction rules.}
  \label{tab:extraction}
\end{table*}

\subsection{Result Analysis}
\begin{table*}[t!]
  \centering
  
    \begin{tabular}{l|cc|cc|cc}
    \toprule
     \backslashbox[48mm]{Methods}{Datasets}&
     \multicolumn{2}{c}{Yelp} & \multicolumn{2}{|c}{IMDB} &
     \multicolumn{2}{|c}{Amazon} \\
     & Mean & Std & Mean & Std & Mean & Std \\
    \midrule
      Lexicon & 0.5982 & 0.0006 & 0.5998 & 0.0013 & 0.5754 & 0.0014 \\
      Keyword Pretraining & 0.7441 & 0.0060 & 0.7496 & 0.0050 & 0.6375 & 0.0171 \\
      WeSTClass \cite{meng2018weakly} & 0.8061 & 0.0105 & 0.7354 & 0.0096 & 0.7374 & 0.0082 \\
      VWS \cite{zeng2019variational} & 0.8014 & 0.0179 & 0.7825 & 0.0045 & 0.7530 & 0.0025 \\
      VWS-PR & \textbf{0.8431} & 0.0013 & \textbf{0.8025} & 0.0041 & \textbf{0.7644} & 0.0020 \\
        \bottomrule
    \end{tabular}%
    \caption{F1 scores of weakly supervised sentiment analysis methods on Yelp, IMDB, and Amazon.}
  \label{tab:result}%
\end{table*}%
Table \ref{tab:result} shows that our method VWS-PR outperforms VWS by $\mathbf{4\%}$, $\mathbf{2\%}$, and $\mathbf{1\%}$ on Yelp, IMDB, and Amazon datasets respsectively. 
Compared with WeSTClass and VWS, our method is more stable, i.e., smaller standard deviation, which shows that the regularization confine the posterior distribution to a smaller space. 
The performance of lexicon method is bad across three datasets. 
The main reason is that it does not involve any learning process. 
Keyword pretraining method can outperform lexicon method. 
But pseudo labels are not ground truths, hence the pseudo training set contains noises. 
Also, user provided keywords are limited, so the training samples with pseudo labels are restricted to some samples which contain certain keywords. For example, reviews with an extreme polarity (only expressing positive polarity or only express negative polarity) are likely to be pseudo samples. 
But most of reviews express mixed polarities.
This will hinder the generalization ability. 
WeSTClass outperforms the keyword pretraining method on Yelp and Amazon dataset due to keyword expansion and self-training process. 
But in IMDB dataset, WeSTClass is slightly worse than the keyword pretraining method. 
Possible reason would be keyword expansion involve some harmful keywords and self-training procedure amplifies errors. 
VWS outperforms WeSTClass on IMDB and Amazon datasets and is comparable to Yelp dataset.

 \subsection{Hyper-parameters Sensitivity Analysis}

\begin{figure*}[t]
	\centering
    \subfigure[Varying $\beta$ with fixed $\gamma_{1}$ and $\gamma_{2}$.]
    {\label{fig:beta}
		\includegraphics[width=0.3\textwidth]{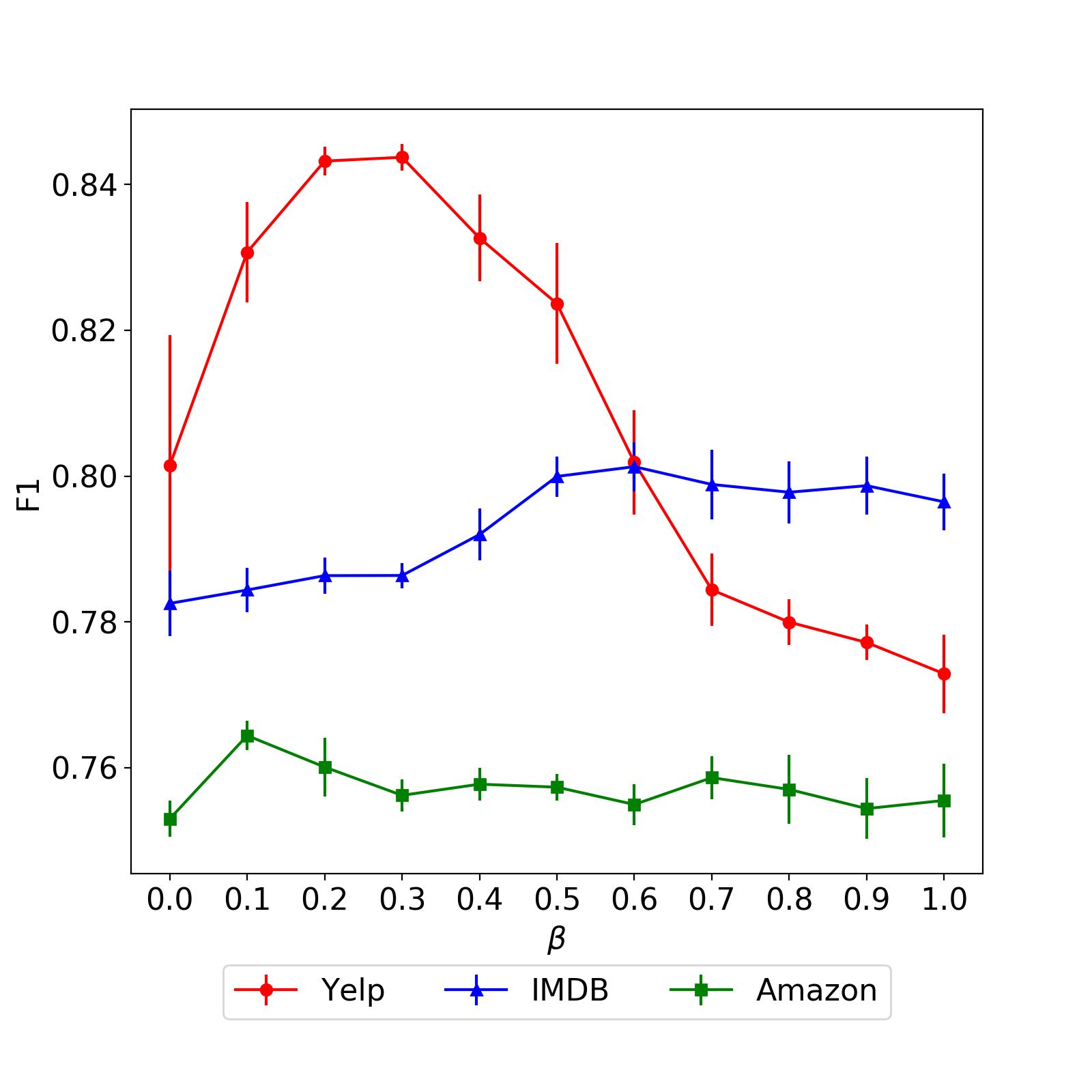}
	}
	\subfigure[Varying $\gamma_{1}$ with fixed $\beta$ and $\gamma_{2}$.]
	{\label{fig:gamma_1}
		\includegraphics[width=0.3\textwidth]{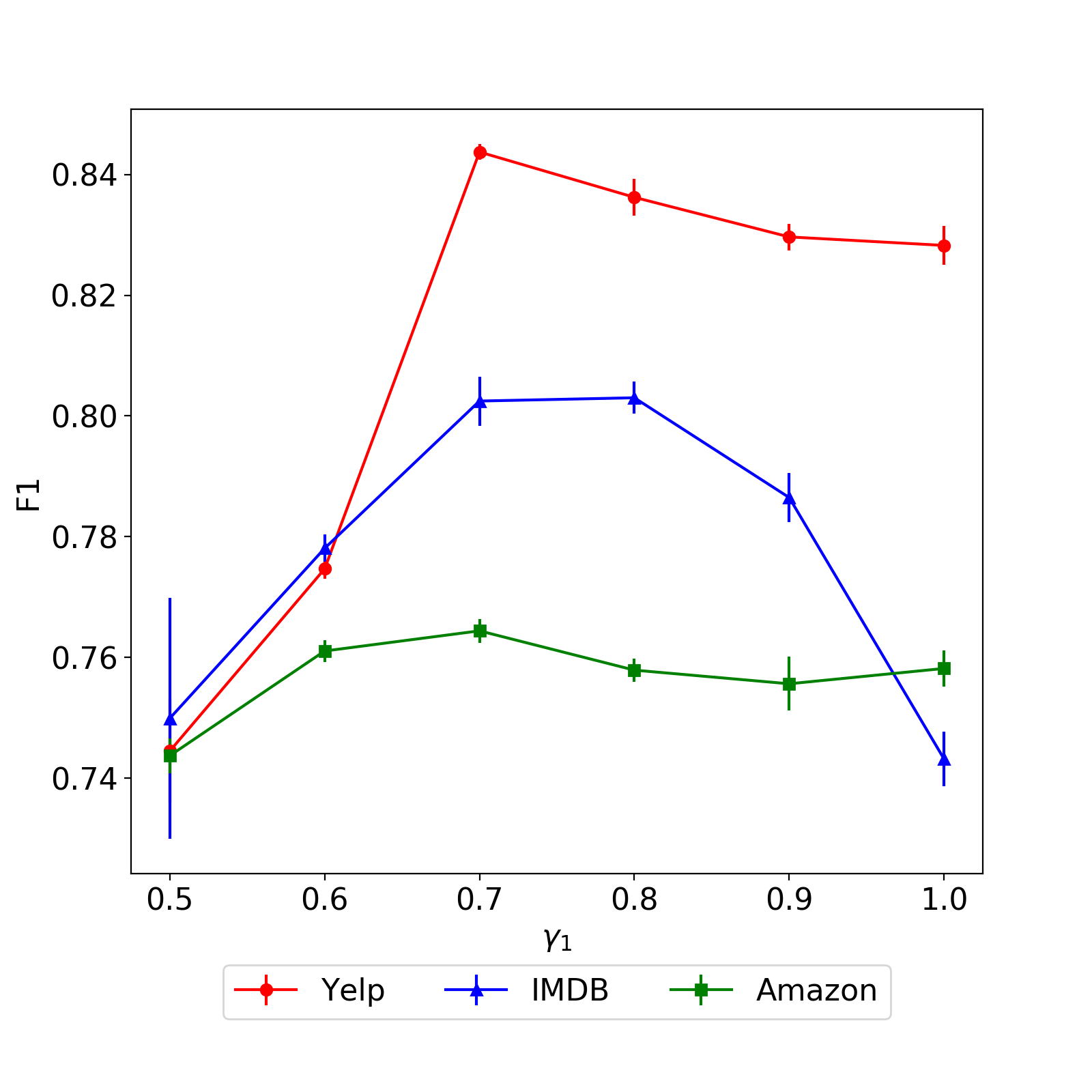}
	}
	\subfigure[Varying $\gamma_{2}$ with fixed $\beta$ and $\gamma_{1}$.]
	{\label{fig:gamma_2}
		\includegraphics[width=0.3\textwidth]{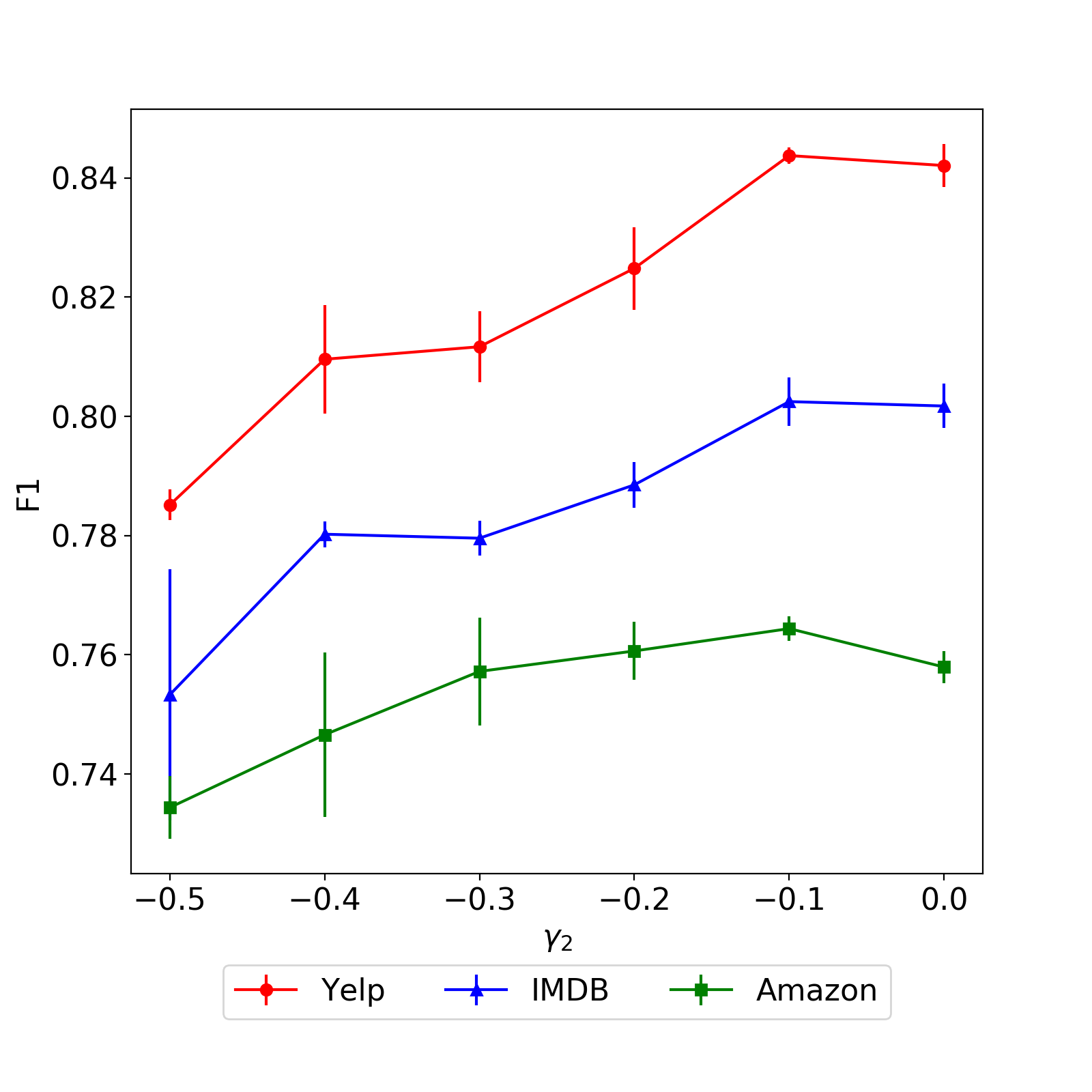}
	}
	\caption{Hyper-parameters sensitivity analysis.  }
    \label{fig:param}
\end{figure*}

We first show F1 scores on three datasets with varied $\beta$ in Figure \ref{fig:beta}. 
It shows that optimal $\beta$ values of our model on three datasets are different. 
When they achieve optimal $\beta$ value, the standard deviation is much smaller than others. The regularization makes models more stable. 
Our method on IMDB and Amazon is not very sensitive. 
The changes are within $2\%$.
Our method on Yelp is more sensitive. 
But we could still find a range, e.g., $0.1$ to $0.5$, where the changes are within $2\%$. 
When $\beta$ keeps growing, the performance in Yelp deteriorates a lot. 
The main reason is probably that the opinion word vocabulary size in Yelp is much larger than other datasets, and hence it is likely that the vocabulary in Yelp contains more noisy opinion words. 
When $\beta$ is large, the noises may harm the performance.

We then show F1 scores on three datasets with varied $\gamma_{1}$ in Figure \ref{fig:gamma_1}. 
The optimal $\gamma_{1}$ values of three datasets are the same, i.e., $0.7$.
The trends are consistent on three datasets. 
F1 score first increases and then decreases.
When $\gamma_{1}$ is small, the constraints are easier to satisfy and the performance is bad because it may involve more noisy constraints. 
It probably enforces two samples to be similar but in reality, they are not similar. 
When $\gamma_{1}$ is large, the performance is also bad because it has fewer constraints that enforce two samples to be similar. 
It could have made use of more constraints. 

We show F1 scores on three datasets with varied $\gamma_{2}$ in Figure \ref{fig:gamma_2}. 
The optimal $\gamma_{2}$ values of three datasets are the same, i.e., $-0.1$.
The trends are consistent on three datasets. 
F1 score first increases and then decreases.
When $\gamma_{2}$ is small, the performance is bad because it has fewer constraints that enforce two samples to be dissimilar. 
It could have made use of more constraints. 
When $\gamma_{2}$ is large, the performance is bad because it has more noisy constraints. 

\subsection{Error Analysis}
\begin{table*}[t!]
  \centering
  \begin{tabular}{l|l|c|c|c}
  \toprule
  Dataset & Document & Prediction & Ground Truth & Extracted Words\\
  \midrule
  Yelp & It was good, but not at that price. & good & bad & good\\
  & There are so many other good italian places & & &  \\
  & in the area for half the price. & & &  \\
  \midrule
  IMDB & I love gheorghe muresan, & bad & good & love \\
  & so i automatically loved this movie. & & &  good \\
  & Everything else about it was so so. & & & annoying \\
  & Billy crystal is a good actor,  & & &  \\
  & even if he is annoying. & & &  \\
  \midrule
  Amazon & No wrist strap. Nice light, well made. & good & bad & nice \\
  & But why would anyone design a & & &  \\
  & tactical style flashlight without & & &  \\
  & a place to attach a wrist strap. & & &  \\
  \bottomrule
  \end{tabular}
  \caption{Documents that are predicted incorrectly by VWS-PR.}
  \label{tab:error}
\end{table*}
We show some incorrectly predicted documents by VWS-PR on three datasets in Table \ref{tab:error}. 
For the first document, the customer emphasizes price a lot. 
Our method cannot extract opinion words on snippets such as ``not at that price'' and ``for half the price.'' 
For the second document, the reviewer loves this movie because he/she loves the basketball player. 
The reviewer thinks that the movie itself does not deserve a high score. 
Our method detects both positive and negative polarities on this document, so it tends to predict as negative. 
Because most mixed polarities are likely to be negative polarity. 
The regularization enforces this pattern.
This document obviously is different from other documents with mixed opinion words. 
For the last document, our method cannot extract opinion words on snippets such as ``no wrist strap'' and ``without a place to attach a wrist strap.'' 
Our method is good at extracting words on subjective expression such as ``nice light,'' but not on descriptive expression such as ``no wrist strap.''
Our method fails because no other knowledge source indicates that ``no wrist strap'' is negative. 
\subsection{Implementation Details}
For WeSTClass and VWS, we used code released by \cite{meng2018weakly} and \cite{zeng2019variational} respectively, and followed their preprocessing steps and optimal settings.
For VWS and VWS-PR, we pretrain a CNN model using pseudo-labeled samples. 
After that, the embeddings are untrainable. 
The rest of parameters are trainable. 
For our method, the hyperparameter settings of VWS part is the same as described in \cite{zeng2019variational}. 
We implemented our models using TensorFlow \cite{abadi2016}.
When tuning hyper-parameters of regularization term, we perform grid search on $\gamma_1 \in [0.5,0.6,\cdots,1.0]$ and $\gamma_2 \in [-0.5,-0.4,\cdots,0]$. After than, we fix $\gamma_1$ and $\gamma2$, then tune $\beta \in [0.1,0.2,\cdots,1.0]$. $\delta$ is fixed to $1$. 
\section{Related Work}
In this section, we review the related work on weakly supervised sentiment analysis.


Using a lexicon is a typical way to perform weakly supervised sentiment analysis. 
One line of works perform simple assignment, i.e., majority voting, based on sentiment orientation scores of extracted opinion words. 
Some methods \cite{missen2009using,tsytsarau2010scalable} used sentiment orientation scores in existing lexicons directly, and aggregated them within a document to determine polarity. 
Some methods developed their own semantic orientation estimation algorithm. 
For example, \cite{turney2002thumbs} first identified phrases in the review and then estimated the semantic orientation of each extracted phrases. 
The semantic orientation of a given phrase is calculated by comparing its similarity to a positive reference word (``excellent'') with its similarity to a negative reference word (``poor'').
This method determined the sentiment polarity based on the average semantic orientation of the phrases extracted from the review. 
\cite{kamps2004using} used the minimum path distance between a phrase and pivot words (``good'' and ``bad'') in WordNet to estimate the semantic orientation of extracted phrases. 

Another line of works involve learning process when using a lexicon.  
\cite{li2009non,zhou2014sentiment} proposed a constrained non-negative matrix tri-factorization approach to sentiment analysis, and used a sentiment lexicon as prior knowledge. 
In these models, a term-document matrix is approximated by three factors that specify soft membership of terms and documents in one of $k$ classes. 
All three factors are non-negative matrices. 
The first factor is a matrix representing knowledge in the word space, i.e., each row represents the posterior probability of a word belonging to the $k$ classes. 
The second factor is a matrix providing a condensed view of the term-document matrix.
The third factor is a matrix representing knowledge in document space, i.e., each row represents the posterior probability of a document belonging to the $k$ classes.
\cite{li2009non} applied a regularization to encourage that the first factor is close to prior knowledge. 
This regularization is different from ours because it requires prior knowledge such as a predefined lexicon. 
A predefined lexicon needs a lot of human effort. 
\cite{zhou2014sentiment} applied a regularization based on an intuition that if two documents are sufficiently close to each other, they tend to share the same sentiment polarity. 
This intuition of this regularization is similar to ours. But when they compare document similarity, they use textual similarity (e.g., cosine similarity of bag of words) rather than similarity on opinion words. The regularization is applied under matrix factorization framework, it is not straightforward to fit in neural network based models. 

Using keywords is another way to perform weakly supervised sentiment analysis. 
\cite{meng2018weakly} leveraged keywords to generate pseudo-labeled samples for model pretraining, and then performed self-training on unlabeled data for model refinement. 
The possible improvement of this direction would be investigating more advanced keywords expansion techniques to generate better pseudo-labeled samples and developing a more advanced self-training algorithm. 
LOTClass \cite{meng2020text}, a parallel work to ours, fine-tuned a masked language model to generate relevant words that can replace label name such as ``good'' and ``bad,'' and performed self-training on unlabeled data for model refinement. 
Fine-tuning in LOTClass can be viewed as an advanced keyword expansion process using language models. 
VWS \cite{zeng2019variational} used target-opinion word pairs as supervision signal. Its objective function is to predict an opinion word given a target word. 
By introducing a latent variable (the sentiment polarity), they can learn a well-approximated posterior distribution via optimizing the evidence lower bound. The posterior probability here is the probability of a possible polarity (e.g., positive or negative) given text representation.

\section{Conclusion}
We propose a posterior regularization framework for the VWS sentiment analysis to better control the posterior distribution. 
The intuition behind the posterior regularization is that if extracted opinion words from two documents are semantically similar (dissimilar), the posterior distribution of two documents should be similar (dissimilar). 
Our experiments show that our posterior regularization 
can improve VWS and the performance is more stable. 
\section*{Acknowledgments}
We thank anonymous reviewers for their valuable comments.
This paper was supported by the NSFC Grant U20B2053 from China, the Early Career Scheme (ECS, No. 26206717),  the General Research Fund (GRF, No. 16211520), and the Research Impact Fund (RIF, No. R6020-19) from the Research Grants Council (RGC) of Hong Kong.

\bibliography{eacl2021}
\bibliographystyle{acl_natbib}

\end{document}